\documentclass[conference]{IEEEtran}
\IEEEoverridecommandlockouts

\usepackage{cite}
\usepackage{amsmath,amssymb,amsfonts}
\usepackage{algorithmic}
\usepackage{graphicx}
\usepackage{textcomp}
\usepackage{xcolor}
\usepackage{hyperref}

\def\BibTeX{{\rm B\kern-.05em{\sc i\kern-.025em b}\kern-.08em
    T\kern-.1667em\lower.7ex\hbox{E}\kern-.125emX}}
\begin{document}

\title{Agentic Control Center for Data Product Optimization\\}

\makeatletter
\newcommand{\linebreakand}{%
  \end{@IEEEauthorhalign}
  \hfill\mbox{}\par
  \mbox{}\hfill\begin{@IEEEauthorhalign}
}
\makeatother

\author{\IEEEauthorblockN{Priyadarshini Tamilselvan}
\IEEEauthorblockA{\textit{Georgia Tech} \\
priya61197@gmail.com}
\and
\IEEEauthorblockN{Gregory Bramble}
\IEEEauthorblockA{\textit{IBM Research} \\
gbramble@us.ibm.com}
\and
\IEEEauthorblockN{Sola Shirai}
\IEEEauthorblockA{\textit{IBM Research} \\
solashirai@ibm.com}
\linebreakand
\IEEEauthorblockN{Ken C. L. Wong}
\IEEEauthorblockA{\textit{IBM Research}\\
clwong@us.ibm.com}
\and
\IEEEauthorblockN{Faisal Chowdhury}
\IEEEauthorblockA{\textit{IBM Research}\\
mchowdh@us.ibm.com}
\and
\IEEEauthorblockN{Horst Samulowitz}
\IEEEauthorblockA{\textit{IBM Research}\\
samulowitz@us.ibm.com}
}

\IEEEpubid{\makebox[\columnwidth]{2025 IEEE International Conference on Data Mining (ICDM) - Demo Track \hfill} \hspace{\columnsep}\makebox[\columnwidth]{ }}

\maketitle
\IEEEpubidadjcol

\begin{abstract}
Data products enable end users to gain greater insights about their data by providing supporting assets, such as example question-SQL pairs which can be answered using the data or views over the database tables. However, producing useful data products is challenging, and typically requires domain experts to hand-craft supporting assets. We propose a system that automates data product improvement through specialized AI agents operating in a continuous optimization loop. By surfacing questions, monitoring multi-dimensional quality metrics, and supporting human-in-the-loop controls, it transforms data into observable and refinable assets that balance automation with trust and oversight.

A video of the demo is available in \href{https://ibm.box.com/s/6kznvxax8j47a4vfjhx287sgix3v8rqv}{\it https://ibm.box.com/s/6kznvxax8j47a4vfjhx287sgix3v8rqv}.
\end{abstract}


\section{Introduction}

\begin{figure}[ht]
\centering
\includegraphics[width=\linewidth]{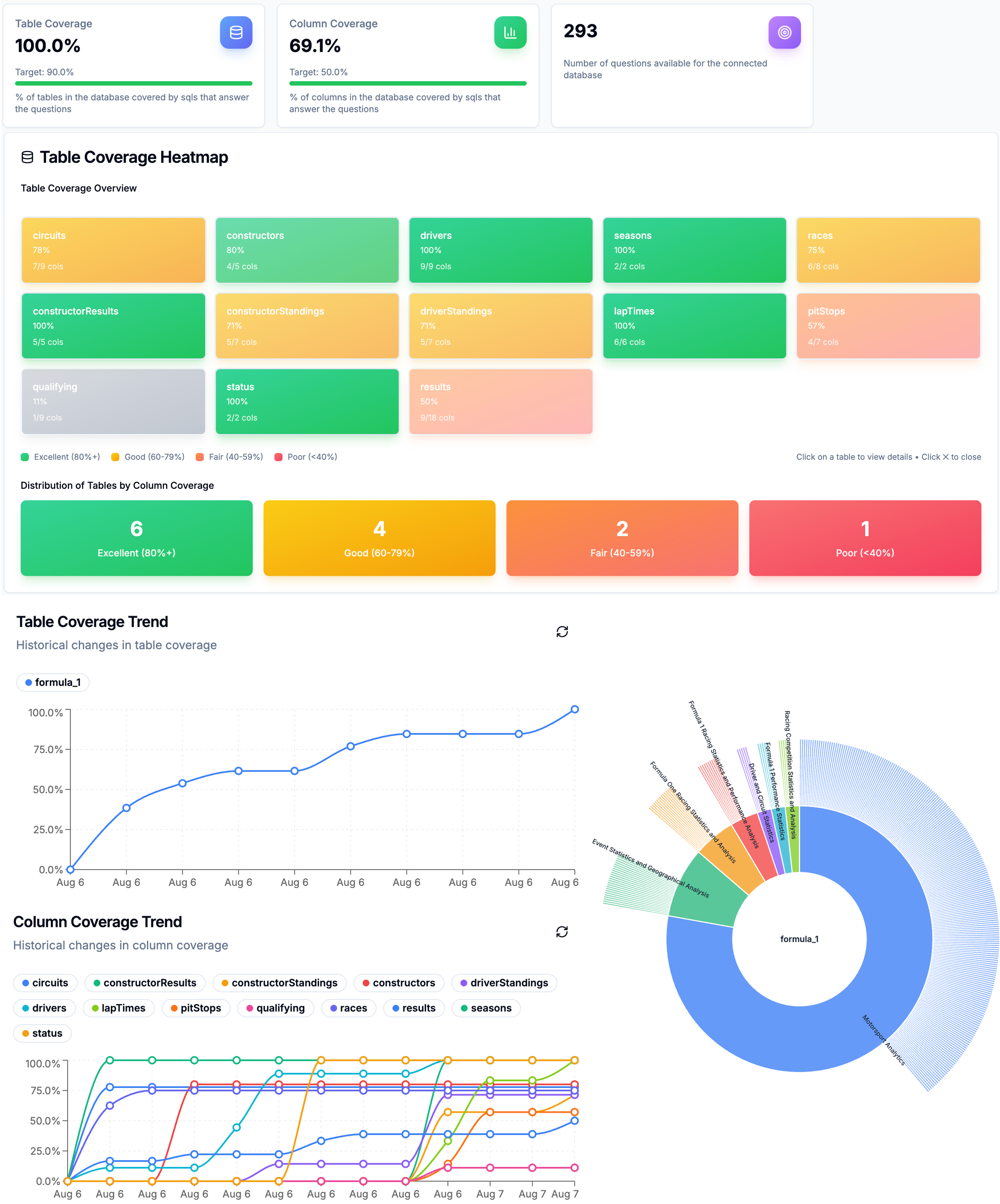}
\caption{The Agentic Control Center allows users to observe key statistics and trends as AI agents perform actions to improve the underlying data product.}
\label{fig:testing}
\end{figure}

As the availability and scope of data collected by organizations continues to grow, it has become increasingly important to \textit{understand} what data you have. To be useful, data relevant to a particular use case must be discoverable. Additionally, having supporting assets for data, such as task-specific views or query examples, can make the use of data more effective. These challenges make it important to move beyond simply having collections of data, but instead creating \textit{data products} to provide greater value.

A data product is a reusable package of data assets for solving business problems or providing new values for customers \cite{IBMdataproduct}. By analyzing the raw data through data engineering, machine learning, and AI   approaches, new insights and added values can be provided in terms of code, metadata, and governing policies \cite{DataProducts}. Therefore, data becomes more insightful when paired with well-designed models, optimized access patterns, and meaningful questions. By surfacing relevant queries, clustering topics, and guiding exploration, raw data is transformed into a living knowledge interface. Traditionally, this has relied on expert data engineers to design views, craft queries, and enforce quality, but this human-centric approach is costly, slow, and difficult to scale.

Large language models (LLMs), especially AI agents, provide new opportunities for automatic data product creation. While some early approaches for benchmarking and evaluation of data products are beginning to emerge \cite{zhangdp2025}, assessing the quality of a data product is nontrivial and can be subjective, and black-box operations of LLMs raise concerns around observability, control, and trust. Hence, visibility of agent actions, ability to provide human feedback, and mechanisms for intervention are desirable.

This paper introduces the Agentic Control Center for Data Product Optimization (Fig. \ref{fig:testing}), a framework that addresses these gaps through the following contributions:
\begin{enumerate}
    \item We present the concept of autonomous data product improvement through measurable quality contracts and optimization objectives.
    \item We demonstrate the benefits of multi-agent collaboration with specialized agents for complex data tasks.
    \item We highlight the importance of human-in-the-loop control mechanisms for production deployment of agentic data systems. 
\end{enumerate}

\section{Related Work}

Prior work has shown the benefits of agentic architectures for use cases such as text-to-SQL. Examples include RAISE~\cite{raise2024}, which is a unified approach for schema linking, SQL generation, and reflection; MAC-SQL~\cite{macsql2023}, which applies specialized agent collaboration; and Tool-Assisted Agent~\cite{toolassisted2024}, which adds quality detection to catch silent errors. In applying agentic approaches to such tasks, human oversight remains critical, as shown by Interactive-KBQA~\cite{interactive2024}. Building upon such ideas, we extend agent specialization to data product generation through the use of planning, execution, and quality check agents, as well as using improvement loops to continuously refine the data product.


Beyond query generation, LLMs are increasingly applied to autonomous insight generation. Data-to-Dashboard~\cite{data2dashboard2024} demonstrates the value of multi-agent pipelines for producing richer visualizations, and QUIS~\cite{quis2024} advances exploratory analysis by refining analytical questions and generating insights from data. Automated data pipelines~\cite{autonomous2024} further articulate LLMs as semi-autonomous analytics partners. Our work advances these approaches by supporting visualizations and optimization based on data quality metrics, supporting both agentic and human-in-the-loop decision making.

The theoretical foundations of multi-agent LLM systems have been studied through frameworks like CAMEL~\cite{camel2023}, which investigated emergent cooperation patterns among AI agents. Its findings on communication and delegation inform our coordination and negotiation between planner and tool agents. NeurDB~\cite{neurdb2024} embeds AI workflows into the database engine, enabling adaptive optimization through storage-layer intelligence and automatic physical design tuning. While NeurDB focuses on storage, our framework operates at the orchestration layer, rewriting SQL, modifying schemas, and improving data quality.

Our system aims to address the gaps in existing work surrounding continuous, contract-aware optimization across the full data lifecycle.

\section{Architecture Overview}
The proposed system implements an autonomous data product optimization framework that continuously improves data quality metrics through intelligent workflow management. The architecture operates on a continuous improvement loop where data product quality is evaluated against explicit user-defined contracts. These contracts serve as the guiding objectives of the system, ensuring that gaps are automatically identified and corrective actions are executed through autonomous agents to uphold the specified quality guarantees.

As shown in Fig. \ref{fig:SystemArchitecture}, the system is built around four core components: a State Manager for data product states, a Data Product Quality Metrics module for defining and tracking measures, a Tool Registry for managing agent integrations, and an Agentic Orchestration Layer for coordinating autonomous optimization workflows. This modular architecture ensures scalability across diverse data environments while maintaining a clear separation of concerns.

\begin{figure}
  \centering
  \includegraphics[width=\linewidth]{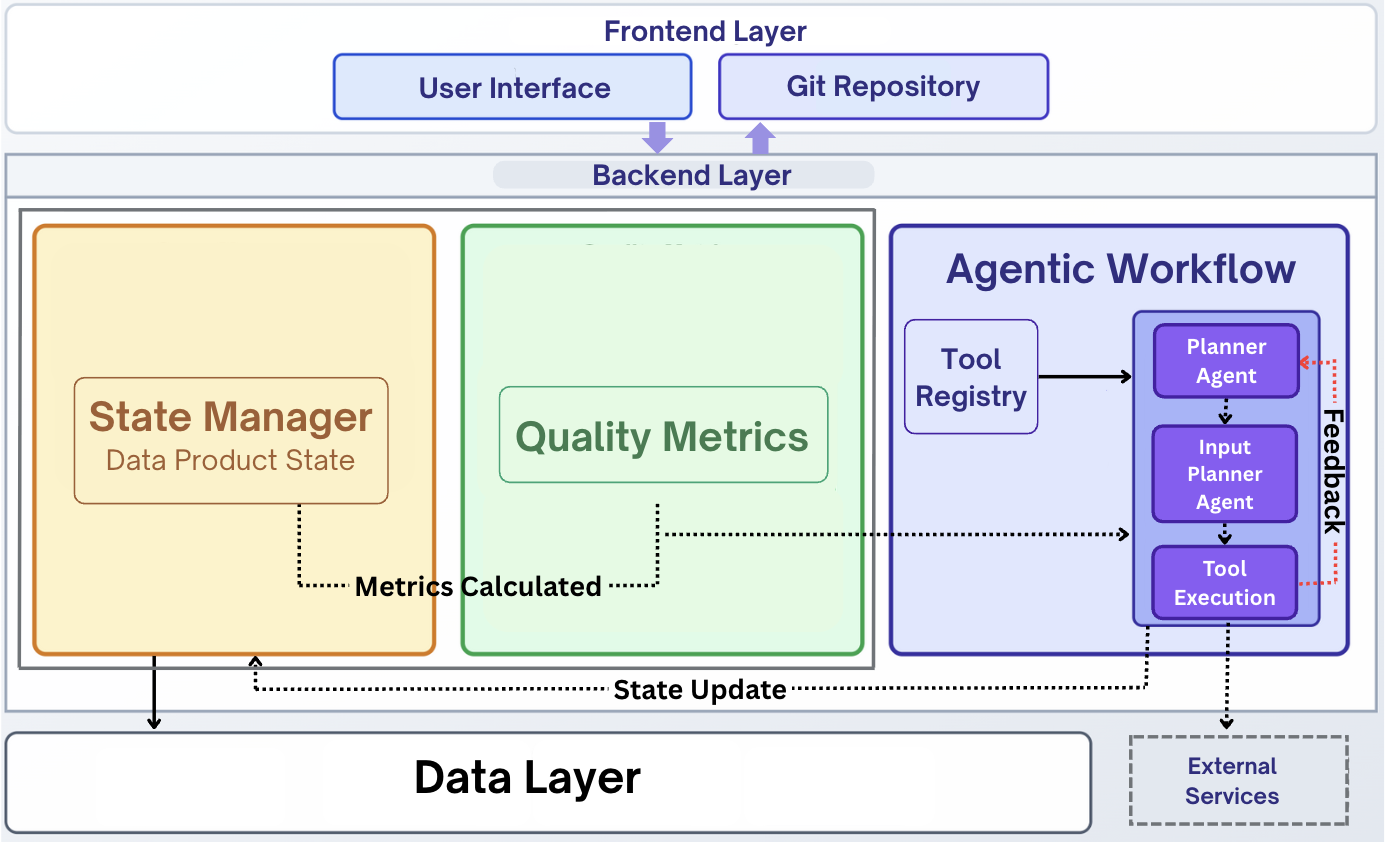}
  \caption{The two-tier Agentic Control Center System Architecture where the frontend connects data sources and sets quality targets. The backend combines a State Manager (tracking state and metrics) with an Agentic Workflow that iteratively plans, executes, and optimizes to achieve those targets.}
  \label{fig:SystemArchitecture}
\end{figure}


\subsection{State Management}

\subsubsection{State Manager}

The state manager serves as the central piece of the architecture, maintaining complete control over the system's operational state. At its core lies the \emph{data product state}, which provides a comprehensive representation of all data products within the system. This state representation drives all metric calculations and serves as the single source of truth for system operations.

\subsubsection{Data Product State}
The data product state encompasses multiple components that together enable comprehensive metric calculation. It maintains \textbf{table} and \textbf{column metadata} to capture schema details from connected data sources, along with \textbf{question mappings} that link predefined questions to schema elements. To support iterative refinement, it tracks \textbf{query versions} which represent the evolutionary history of SQL associated with each question, and \textbf{answer versioning} which records returned answers with confidence scores for performance monitoring. 
In addition, the state stores \textbf{view definitions} used in query optimization.

When data sources are connected, the State Manager populates schemas, tables, and columns through a flexible abstraction layer, supporting databases like SQLite, MySQL, PostgreSQL, and BigQuery without code changes. Supplementary metadata, such as predefined questions, can be incorporated at this stage, enabling the system to build on existing inputs rather than starting from scratch.

\subsection{Data Product Quality Metrics Management}
\subsubsection{Data Product Quality Metrics and Dependencies}
The system implements a configurable approach to defining data product quality metrics, the specific measurements that quantify aspects like table and column coverage, query speed and complexity, and response accuracy. These  metrics are registered with the system through a dynamic definition mechanism that allows users to specify what \textit{improvement} means for their particular data products.
The metric management system handles the dependencies between different quality measurements, ensuring that when one aspect of the data product changes, all dependent metrics are appropriately recalculated.

\begin{figure}
  \centering
  \includegraphics[width=\linewidth]{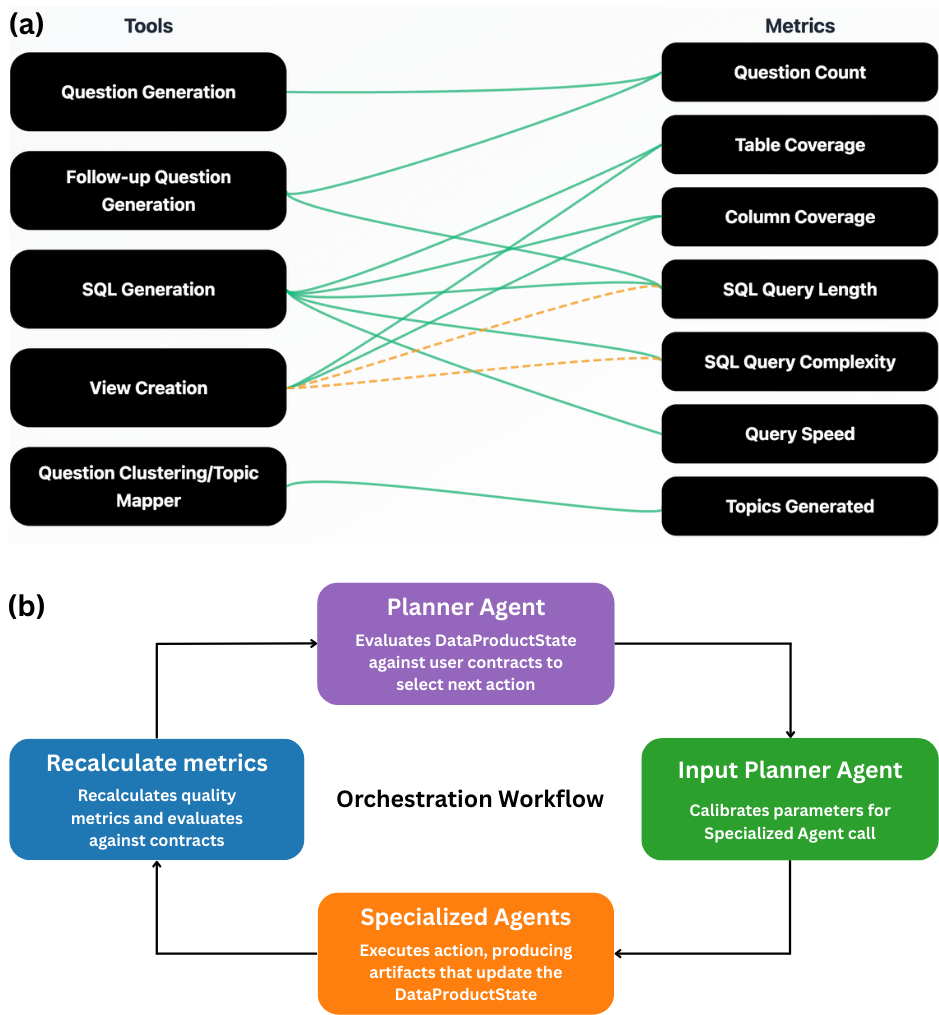}
  \caption{(a) Mapping of tools (e.g., question generation,  view creation) to quality metrics such as coverage, complexity, and speed, with arrows showing positive (green) or optimizing (orange) effects. (b) The Planner Agent selects actions, the Input Planner Agent calibrates parameters, Specialized Agents execute updates, and metrics are recalculated for iterative optimization.}
  \label{fig:ToolImpactAndLoop}
\end{figure}

\subsubsection{Context Generation and Event-Driven Recalculation}
The system implements context generation logic to determine appropriate calculation scopes upon state changes. When components are added, modified, or removed, events propagate through the metric dependency graph. The context generator analyzes these events to identify which metrics require recalculation and at which granularity. For example, when a new table is added, the system recalculates table coverage metrics at the database level and column coverage metrics at the table level. This hierarchical context resolution minimizes unnecessary recalculation while ensuring metric consistency.

\subsection{Tool Management and Integration}

The Tool Registry provides a flexible mechanism for registering and discovering external tools. Any tool can be added by specifying its name, description, input parameters, output schema, and execution context. In practice, tools may correspond to agents or other services, and can be invoked dynamically based on system state and task requirements. As shown in Fig. \ref{fig:ToolImpactAndLoop}(a), these tools are linked to specific quality metrics, illustrating how their outputs drive measurable improvements across the system. Question and Follow-up Generation increase question count, with the latter also lengthening SQL queries. SQL Generation improves table and column coverage, while influencing query length, complexity, and speed. View Creation reduces query length and complexity but expands coverage, and Topic Mapping clusters queries into structured topics. Together, these components balance query richness, coverage, and efficiency.

\section{AI Agents and Orchestration}

The orchestration layer implements autonomous workflow management through specialized AI agents that work in concert to optimize data product metrics against user-defined contracts. The overall workflow is illustrated in Fig.~\ref{fig:ToolImpactAndLoop}(b), which depicts how agents coordinate to plan, execute, and refine optimization tasks in iterative cycles.

\subsubsection{Planner Agent}
The Planner Agent serves as the system's central decision-maker, continuously evaluates the DataProductState against completeness and performance contracts. It analyzes actual metric values versus targets using a tool precondition graph to identify the single highest-impact action. The agent outputs a structured JSON payload containing the selected tool name, target context identifiers (database/table/question), expected improvement magnitude, and a human-readable rationale. For example, detecting low table coverage triggers the call to the Question Generator agent to address the gap.

\subsubsection{Input Planner Agent}
The Input Planner Agent transforms high-level action requests into precisely calibrated tool parameters by analyzing schema characteristics, historical outputs, and contract gaps. It employs adaptive heuristics to tailor the scale of generation to the state of the system. For example, it may produce 80 questions when more than 50 tables remain under-covered, while limiting this to around 20 when only a handful of tables require attention, and it similarly adjusts the number of views generated to support performance tuning. This intelligent parameterization ensures each tool invocation maximizes impact while maintaining computational efficiency for hyperparameter optimization.

\subsubsection{Specialized Agents}
The Specialized Agents constitute modular components registered within the system, each responsible for producing service-specific artifacts that collectively advance the state of the data product. In the current implementation, the following specialized agents are available: (i) Question Generation Agent, (ii) Text-to-SQL Agent, (iii) Follow-up Question Generation Agent, (iv) Question Clustering Agent, and (v) View Creation Agent. The generated artifacts can include question lists, SQL statements, cluster assignments, and views. Each tool specializes in a different aspect of the pipeline: question generation using QUIS~\cite{quis2024} introduces novel queries and associated metadata, text-to-SQL tools produce executable SQL statements and query plans, view creation agents generate SQL views, follow-up generators construct related or chained questions to enrich exploration, and clustering agents group questions to improve organization and coverage.  Post-processing consolidates these heterogeneous artifacts into the unified \textit{DataProductState}, ensuring consistency and traceability across iterations. 

\subsubsection{Recalculate Metrics}
After the specialized agents are run, any affected quality metrics (e.g., \textit{coverage, query length and complexity, execution speed}) are recalculated to reflect the updated state. Additionally, the resulting outputs, most notably SQL queries and derived views, are committed to an integrated Git repository for version control and auditability.

The system implements a continuous improvement loop where Tool Agents return control to the Planner Agent after each execution, once metrics are re-profiled. This cycle of \texttt{plan → parameterize → execute → update → measure} continues until contractual objectives are met or no further improvements are possible. As illustrated in Fig. \ref{fig:ToolImpactAndLoop}(b), this separation of strategic planning, tactical parameterization, and operational execution enables sophisticated autonomous optimization while preserving modularity through well-defined interfaces between agents.

\section{Case Study}


To analyze the effectiveness of the \textit{Agentic Control Center}'s autonomous orchestration capabilities, we conducted a case study using three representative databases from the BIRD~\cite{BIRDbench} benchmark.
Each database was configured with quality targets: 90\% table coverage, 50\% column coverage, and performance thresholds requiring average query execution times under 5 seconds. Our three databases provided a range of scales and complexity profiles surrounding the number of tables, data types, and relational structure.

We observe that our system demonstrated markedly different optimization strategies based on database characteristics. In our smallest database, the system exhibited rapid convergence, quickly identifying coverage gaps and satisfying quality targets within a few optimization iterations. Conversely, the more complex database triggered sophisticated adaptive strategies. The system demonstrated intelligent parameter synthesis by automatically increasing question generation quantities and selecting tools for more complex and multi-stage question generation. This in turn facilitated generation of more complex SQL queries (e.g., multi-level subqueries and joins). 

These behaviors were also reflected in the Input Planner agent's logs, which revealed how the agent could systematically prioritize high-impact actions such as targeting large, unexercised tables which would be expected to yield maximum coverage improvements at each iteration. When encountering databases with numerous small tables, the system automatically adjusted strategies to surface complex join relationships rather than limiting exploration to simple lookups. 

Additionally, we observed that the Planner Agent was capable of detecting diminishing returns and transitioning between stages at optimal points. In cases where further autonomous actions yielded minimal coverage improvements, the system appropriately recommended manual review rather than continuing inefficient iterations, indicating sophisticated meta-reasoning about its own effectiveness and limitations.

Once coverage objectives were satisfied, the question clustering tool addressed the challenge of understanding hundreds of generated questions by grouping related queries into coherent topics, providing users with a high-level overview of the kinds of questions that could be answered using the data product.
The integrated Git workflow provided comprehensive auditability throughout the validation process, with every autonomous decision, generated artifact, and human modification automatically versioned. This controlled modification approach prevented arbitrary changes. 

\section{Conclusion}
We introduced an LLM agent-driven system for optimizing data products via a transparent and dynamic user interface. It leverages metrics derived from data tools to guide optimization of an evolving data product. This prototype represents an initial step toward a comprehensive data optimization framework, which will require an expanded set of metrics, additional data tools, a more scalable and interactive interface, and the ability to solve complex multi-objective optimization problems.

\end{document}